# The JRC-Acquis: A Multilingual Aligned Parallel Corpus with 20+ Languages

## Ralf Steinberger[1], Bruno Pouliquen[1], Anna Widiger[1], Camelia Ignat[1]
## Tomaž Erjavec[2], Dan Tufiş[3], Dániel Varga[4]

(1) European Commission – Joint Research Centre – 21020 Ispra (VA) – Italy
(2) Jožef Stefan Institute – Department of Knowledge Technologies – 1000 Ljubljana – Slovenia
(3) Romanian Academy of Sciences – Research Institute for AI – Bucharest 050711 – Romania
(4) Budapest University of Technology and Economics – Media Research Centre – 1111 Budapest – Hungary
E-mail: (1) Firstname.Lastname@jrc.it, (2) tomaz.erjavec@ijs.si, (3) tufis@racai.ro, (4) daniel@mokk.bme.hu

## Abstract

We present a new, unique and freely available parallel corpus containing European Union (EU) documents of mostly legal nature. It is available in all 20 official EU languages, with additional documents being available in the languages of the EU candidate countries. The corpus consists of almost 8,000 documents per language, with an average size of nearly 9 million words per language. Pair-wise paragraph alignment information produced by two different aligners (Vanilla and HunAlign) is available for all 190+ language pair combinations. Most texts have been manually classified according to the EUROVOC subject domains so that the collection can also be used to train and test multi-label classification algorithms and keyword-assignment software. The corpus is encoded in XML, according to the Text Encoding Initiative Guidelines. Due to the large number of parallel texts in many languages, the JRC-Acquis is particularly suitable to carry out all types of cross-language research, as well as to test and benchmark text analysis software across different languages (for instance for alignment, sentence splitting and term extraction).

## 1. Introduction – Motivation

Parallel corpora are widely sought after, for instance (a) to train automatic systems for Statistical Machine Translation (e.g. Koehn 2005) or multilingual categorisation, (b) to produce multilingual lexical or semantic resources such as dictionaries or Ontologies (e.g. Giguet & Luquet 2005), and (c) to train and test multilingual information extraction software. Additional application areas would be (d) automatic translation consistency checking and (e) the training of multilingual subject domain classifiers (e.g. Pouliquen et al. 2003; Civera & Juan 2006). Due to the many languages of the corpus, we suggest that it is furthermore useful (f) to test and benchmark sentence (and other) alignment software because such software may perform unevenly well for different language pairs.

Most available parallel corpora exist for a small number of languages and mainly involving at least one widely-spoken language, such as the French-English Hansards (Germann 2001) or the English-Norwegian parallel corpus ENPC (http://www.hf.uio.no/ilos/forskning/forskningsprosjekter/enpc). Parallel corpora in more languages are available either for small amounts of text and/or for very specialised texts, such as the bible (Resnik et al. 1999) or the novel *1984* by George Orwell (Erjavec 2004). To our knowledge, the currently most multilingual corpus with a considerable size and variety is *EuroParl* (Koehn 2005), which exists in eleven European languages. EuroParl is offered with bilingual alignments in all language pairs involving English. However, this corpus does not contain any of the languages of the new Member States or of a candidate country.

To our knowledge, the corpus presented in this paper, the *JRC-Acquis* with its 20+ European languages is the most multilingual parallel corpus of its size currently in existence. It is available in TEI-compliant XML format, and includes marked-up texts and bilingual alignment information for all the 190+ language pairs, including rare language combinations such as Estonian-Greek and Maltese-Danish.

An additional feature of the JRC-Acquis is the fact that most texts have been manually classified into subject domains according to the EUROVOC thesaurus (EUROVOC 1995), which is a classification system with over 6000 hierarchically organised classes. Knowing the subject domain(s) of texts can be exploited to produce domain-specific terminology lists, as well as to test and train document classification software and automatic indexing systems. Due to the combination of multi-linguality and subject domain coding of the JRC-Acquis, such systems cannot only be trained multi-monolingually for 20+ languages (e.g. Pouliquen et al. 2003), but new approaches can be developed that exploit data from more than one language at a time (Civera & Juan, 2006).

Exploiting such multilingual corpora has been at the heart of the European Commission's (EC) *Joint Research Centre* (JRC) for years, but we were not allowed to share our textual resources. Now, the EC's *Office for Official Publications* OPOCE has given its agreement that at least the JRC-Acquis collection can be made freely available for research purposes.

The main interest of the JRC in exploiting this highly multilingual parallel corpus stems from the fact that it includes the 'new' EU languages. For some of these, only few linguistic resources are available. In ongoing work, we are attempting to extract general and domain-specific terminology lists and to align these terminology lists across languages to produce multilingual term dictionaries. We aim to use these resources to link similar texts across languages (Steinberger et al., 2004), to improve further the automatic multilingual and cross-lingual news analysis system NewsExplorer[1] (Steinberger et al. 2005), and to offer cross-lingual glossing applications, i.e. to identify known terms in foreign language texts and to display these terms to the users in their own language.

---

[1] Accessible at http://press.jrc.it/NewsExplorer.

In the next sections we explain what the JRC-Acquis is (Section 2), how we compiled it (3) and converted it into clean UTF-8 encoded XML texts with paragraph marking (4). We then summarise the effort to paragraph-align the JRC-Acquis (5), using two alternative approaches, describe the format of the corpus and the alignment information (6) and explain possible uses of the EUROVOC subject domain classification of the corpus (7). Section 8 summarises the work.

## 2. What is the JRC Collection of the *Acquis Communautaire* (Short: the JRC-Acquis)

*EU/EC Acquis Communautaire* is the French and most widely used term to name the body of common rights and obligations which bind all the Member States together within the European Union (EU) (formerly European Community – EC). We will refer to this collection as the *AC* or the *Acquis*. The Acquis is constantly evolving and comprises: the contents, principles and political objectives of the Treaties; EU legislation; declarations and resolutions; international agreements; acts and common objectives. Countries wanting to join the EU have to accept and adopt the Acquis. By definition, translations of this document collection are therefore available in all twenty official EU languages, and translations into Croatian, Romanian, Bulgarian and Turkish are in preparation. The current corpus version contains texts in all 20 official EU languages plus Romanian. We hope to add documents in the languages of the other candidate countries soon.

While a defining list of AC documents should theoretically exist, we have not been able to get hold of this, so that we had to infer which documents available on the EU and other web sites are part of the collection. We decided to select all those documents which exist in at least ten of the twenty-one languages and which are available for at least three of the languages of the Member States who joined the EU in 2004 or in Romanian. As the corpus we compiled is not exactly identical with the legally binding document collection, we use the term *JRC Collection of the Acquis Communautaire* (short: *JRC-Acquis*) to refer to the documents contained in our corpus.

The corpus, related alignment information and documentation are freely available for research purposes and can be downloaded from http://wt.jrc.it/lt/Acquis/. For commercial or other uses, please contact opoce-info-copyright@cec.eu.int.

Like most other official documents of the European Commission and the European Parliament, the Acquis texts have been classified according to the multilingual, hierarchically organised EUROVOC (1995) thesaurus. The main subject domains assigned to the document collection, listed in Table 1, show that the texts cover various subject domains, including economy, health, information technology, law, agriculture, food, politics and more.

## 3. Compiling the JRC-Acquis

Most EU documents are uniquely identifiable by their CELEX code, which consists of a one-digit document type, four-digits to express the year, one letter, four digits and – optionally – brackets containing a one or two-digit number. An example for a decision that entered into force

| IMPORT | EC COUNTRIES |
|---|---|
| PREVENTION OF DISEASE | AGRICULTURAL PRODUCT |
| ORIGINATING PRODUCT | TARIFF QUOTA |
| THIRD COUNTRY | VETERINARY INSPECTION |
| HEALTH CERTIFICATE | FOODSTUFF |
| MARKETING STANDARD | AMNESTY INTERNATIONAL |
| INFORMATION TRANSFER | ANIMAL PRODUCT |
| MARKETING | PLO |
| APPROXIMATION OF LAWS | FISHERIES PRODUCT |

Table 1. Most frequently used EUROVOC descriptors in the JRC-Acquis collection, indicating the most important subject domains of the JRC-Acquis.

in 1999 is 21999D0624(01). The translations of each document have the same unique CELEX identifier.

As of February 2006, most documents in the official EU languages could be found in html format on the Commission's web site http://europa.eu.int/, using the following URL, where the two parameters CELEXCODE and LG should be replaced by their respective values for the Celex code and the two-digit language code:
http://europa.eu.int/smartapi/cgi/sga_doc?smartapi!celexplus!prod!CELEXnumdoc&numdoc=CELEXCODE&lg=LG.
Whenever Celex codes contain brackets, all documents with the same base number (but with different numbers in brackets) are listed on the same page and can be downloaded with the following URL:
http://europa.eu.int/eur-lex/lex/LexUriServ/LexUriServ.do?uri=CELEX:CELEXCODE:LG:HTML.

Documents in the languages of the candidate countries (Romania, Bulgaria, Croatia and Turkey) for which a translation exists, are currently only available in Microsoft Word format. In September 2005, we downloaded the Romanian texts of the JRC-Acquis, using the URL:
http://ccvista.taiex.be/Fulcrum/CCVista/$LG/$CELEXCODE-$LG.doc. The Romanian documents were converted from their original Microsoft Word format to the xml format of the JRC-Acquis corpus. During the automatic conversion, the annotations of the translators and some of the footnotes were discarded. Documents on the ccvista-server do not yet have official status and the translations may still change. We have not yet attempted to download texts in the other candidate country languages.

For some reason, not all language versions are available for all AC documents, and some documents have a non-English title but the text body is in English, and vice-versa. The size of the current version of the JRC-Acquis collection for the various languages is shown in Table 2.

## 4. Pre-processing the JRC-Acquis

After having crawled the mentioned EC web sites and downloaded the selected HTML documents, we converted them to UTF-8-encoded XML format, verified the language using an n-gram-based in-house language guessing software and discarded those documents that were not in the expected language.

Each document was then split into numbered paragraph chunks, using the <BR> or <P> tags from the original HTML documents. As the Acquis texts are consistent and well-structured, these paragraph chunks are

```xml
<?xml version="1.0" encoding="utf-8" ?>
<!DOCTYPE TEI.2 (View Source for full doctype...)>
- <TEI.2 id="jrc42004D0097-fr" n="42004D0097" lang="fr">
  - <teiHeader lang="en" date.created="2006-03-05">
    - <fileDesc>
      - <titleStmt>
          <title>JRC-ACQUIS 42004D0097 French</title>
          <title>2004/97/CE,Euratom: Décision prise du commun accord des représentants des États membres réunis au
            niveau des chefs d'État ou de gouvernement du 13 décembre 2003 relative à la fixation des sièges de
            certains organismes de l'Union européenne</title>
        </titleStmt>
        <extent>40 paragraph segments</extent>
      - <publicationStmt>
        - <distributor>
            <xref url="http://wt.jrc.it/lt/acquis/*">http://wt.jrc.it/lt/acquis/</xref>
          </distributor>
        </publicationStmt>
      - <notesStmt>
          <note>Only European Community legislation printed in the paper edition of the Official Journal of the European
            Union is deemed authentic.</note>
        </notesStmt>
      - <sourceDesc>
        - <bibl>
            Downloaded from
            <xref url="http://europa.eu.int/eur-lex/lex/LexUriServ/LexUriServ.do?
              uri=CELEX:42004D0097:fr:HTML">http://europa.eu.int/eur-lex/lex/LexUriServ/LexUriServ.do?
              uri=CELEX:42004D0097:fr:HTML</xref>
            on
            <date>2006-02-20/22</date>
          </bibl>
        </sourceDesc>
      </fileDesc>
    - <profileDesc>
      - <textClass>
          <classCode scheme="eurovoc">4180</classCode>
          <classCode scheme="eurovoc">5769</classCode>
        </textClass>
      </profileDesc>
    </teiHeader>
  - <text>
    - <body>
        <head n="1">2004/97/CE,Euratom: Décision prise du commun accord des représentants des États membres réunis
          au niveau des chefs d'État ou de gouvernement du 13 décembre 2003 relative à la fixation des sièges de certains
          organismes de l'Union européenne</head>
      - <div type="body">
          <p n="2">Décision prise du commun accord des représentants des États membres réunis au niveau des chefs
            d'État ou de gouvernement</p>
          <p n="3">du 13 décembre 2003</p>
          <p n="4">relative à la fixation des sièges de certains organismes de l'Union européenne</p>
          <p n="5">(2004/97/CE, Euratom)</p>
          <p n="6">LES REPRÉSENTANTS DES ÉTATS MEMBRES, RÉUNIS AU NIVEAU DES CHEFS D'ÉTAT OU DE
            GOUVERNEMENT,</p>
          <p n="7">vu le traité instituant la Communauté européenne, et notamment son article 289, et le traité
            instituant la Communauté européenne de l'énergie atomique, et notamment son article 189,</p>
          <p n="8">considérant ce qui suit:</p>
```

Figure 1: Sample of the TEI header and of the first few lines of a French JRC-Acquis document in XML format.

mostly the same across languages. Each of these paragraphs can contain a small number of sentences, but they sometimes contain sentence parts (ending with a semicolon or a comma) because legal documents frequently

```xml
<div type="signature">
  <p n="20">Done at Brussels, 21 December 1984.</p>
  <p n="21">For the Commission</p>
  <p n="22">Karl-Heinz NARJES</p>
  <p n="23">Member of the Commission</p>
  <p n="24">(1) OJ No 196, 16. 8. 1967, p. 1.</p>
  <p n="25">(2) OJ No L 259, 15. 10. 1979, p. 10.</p>
</div>
```

Figure 2. Typical signature of JRC-Acquis document.

specify their scope with a single sentence spanning over several paragraphs. For an example see Figure 1. As a result, each paragraph of the text collection can be uniquely identified using the language, the CELEX identifier and the paragraph number.

The main *body* of the Acquis texts frequently ends with place and date of *signature* of the document, lists of person names and references to other documents (see Figure 2). Approximately half of the documents furthermore contain an *annex*, which can consist of plain text, lists of addresses, lists of goods, etc. In order to allow users to easily make use of the different sections, we attempted to identify these sections using regular expressions, and mark them up as body, signature and annex (see Figures 1 and 2). This division into three document parts allows users to concentrate their effort on the text type that is most useful for them: While the text body, for instance, rather reliably contains text, the signatures (which are frequently multilingual) contain many named entities (persons, places, dates, references to other documents) so that they could be a good object for named entity recognition tasks. Note that signatures and annexes are usually marked up, but as they were not always clearly identifiable, we will have missed the mark-up of some of them.

| Language | N° of Texts | Text body | | | Signature | Annex | Total N° Words (Text + Signature + Annex) |
|---|---|---|---|---|---|---|---|
| | | Total N° Words | Total N° Characters | Average N° Words | Total N° Words | Total N° Words | |
| cs | **7983** | 6000751 | 38625616 | 751 | 715895 | 1972356 | 8689002 |
| da | 7939 | 6556131 | 44497890 | 825 | 778125 | 1505554 | 8839810 |
| de | 7913 | 6481949 | 46628367 | 819 | 822797 | 1349791 | 8654537 |
| el | 7782 | 7267113 | 47260657 | 933 | 991962 | 1306164 | 9565239 |
| en | 7972 | 7547154 | 45372451 | 946 | 817085 | 1568297 | 9932536 |
| es | 7809 | **8006579** | 48547661 | **1025** | 792355 | 1707348 | **10506282** |
| et | 7943 | 4998334 | 39077676 | **629** | 431570 | 1752216 | 7182120 |
| fi | 7735 | 5141742 | 43771107 | 664 | 618042 | 1120613 | 6880397 |
| fr | 7862 | 7814912 | 46526758 | 994 | 865693 | 1531754 | 10212359 |
| hu | 7489 | 5403934 | 40697500 | 721 | 594176 | 1821141 | 7819251 |
| it | 7872 | 7305910 | 47068517 | 928 | 787131 | 1582773 | 9675814 |
| lt | 7965 | 5395807 | 40011655 | 677 | 693712 | 1869631 | 7959150 |
| lv | 7980 | 5513265 | 38544761 | 690 | 669890 | 1946340 | 8129495 |
| mt | 7639 | 7273072 | 44168004 | 952 | 588103 | 2162699 | 10023874 |
| nl | 7882 | 7362017 | 47846520 | 934 | 806123 | 1593619 | 9761759 |
| pl | 7968 | 6002780 | 43373027 | 753 | 757757 | 1953001 | 8713538 |
| pt | 7848 | 7900690 | 47529143 | 1006 | 746070 | 1692161 | 10338921 |
| ro | 5792 | 5125673 | 33701702 | 884 | 474159 | 3972847 | 9572679 |
| sk | **5278** | **3914177** | 26091927 | 741 | 510522 | 1282176 | **5706875** |
| sl | **7983** | 5954252 | 37646679 | 745 | 743699 | 2017049 | 8715000 |
| sv | 7731 | 6797710 | 44589319 | 879 | 303527 | 1356547 | 8457784 |
| Average | **7,636** | **6.369,712** | **42,456,045** | **833** | **690,876** | **1,764,956** | **8,825,544** |

Table 2: Size of the JRC collection of the Acquis Communautaire in each of the 20 official languages of the European Union and in Romanian. Numbers are given separately for the text body (the main text), the signature and the annexes.

## 5. Paragraph Alignment for all 190 Language Pair Combinations

For the paragraph alignment, we used two different tools to align all texts for all 210 language pair combinations: Vanilla (available at http://nl.ijs.si/telri/Vanilla/), which implements the Gale & Church (1993) alignment algorithm; and HunAlign (Varga et al. 2005). The results for the alignments are available with the distribution of the corpus so that users can use the alignment that suits them best, or for benchmarking exercises. We have not yet been able to carry out a comparative quantitative evaluation of the performance of both tools.

### 5.1. Alignment Using Vanilla

Vanilla is a purely statistical aligner which bases its alignment guesses exclusively on sentence length. As input, it takes parallel texts segmented by hard (reliable) and soft (probable) links. For lack of time, we have not made use of hard links inside the JRC-Acquis documents, although the documents contain numbered sections and paragraphs which could well serve as hard links. Instead, we used soft links directly for paragraph breaks.

Vanilla aligns according to the arities 1-1 (one-to-one), 1-2 (splitting), 2-1 (combination), 1-0 (sentence deletion), 0-1 (sentence insertion) and 2-2. As an average for all language pairs, 85% of the paragraphs of the JRC-Acquis collection was aligned 1-1, which is roughly in line with the sentence alignment results of 89% reported by Church & Gale (1993). The average number of paragraph alignments per language pair is 269,148.

### 5.2. Alignment Using HunAlign

We also processed the corpus using HunAlign, a language-independent sentence aligner (Varga et al, 2005). Unlike Vanilla, HunAlign does not emit 2-2 segments, but it can deal with the splitting of a sentence into more than two sentences. For a fixed choice of language pair, the HunAlign algorithm runs in three phases.

First, it builds alignments using a simple similarity measure. This measure is based on sentence lengths and the ratio of identical words. Number tokens are treated specially: similarity of the sets of number tokens in the two sentences is considered. This special treatment is

```xml
<?xml version="1.0" encoding="utf-8" ?>
<!DOCTYPE div (View Source for full doctype...)>
- <div type="body" n="31960D0511" select="et mt" id="jrc31960D0511-et-mt" org="uniform" sample="complete" part="N"
    TEIform="div">
    <head lang="et" n="1" TEIform="head">Euroopa Ühenduste Komisjon: Otsus, millega kinnitatakse Euratomi tarneagentuuri
      ülesannete alguskuupäev ning kiidetakse heaks tarneagentuuri 5. mai 1960. aasta eeskiri, millega nähakse ette, kuidas
      maakide, lähtematerjalide ja lõhustuvate erimaterjalide nõudlust pakkumisega tasakaalustada</head>
    <head lang="mt" n="1" TEIform="head">Id-DEĊIŻJONI li tiffissa d-data meta l-Aġenzija għall-Forniment tal-Euratom għandha
      tibda taqdi d-dmirijiet tagħha u li tapprova r-Regoli ta' l-Aġenzija tal-5 ta' Mejju 1960 li jistabbilixxu l-manjiera li fiha d-
      domanda għandha tiġi bilanċjata kontra l-provvista ta minerali, materjali primi u materjali speċjali fissili</head>
  - <ab type="1-1" part="N" TEIform="ab">
      <seg lang="et" n="2" part="N" TEIform="seg">Otsus,</seg>
      <seg lang="mt" n="2" part="N" TEIform="seg">Id-deċiżjoni</seg>
    </ab>
  - <ab type="1-1" part="N" TEIform="ab">
      <seg lang="et" n="3" part="N" TEIform="seg">millega kinnitatakse Euratomi Tarneagentuuri ülesannete alguskuupäev ning
        kiidetakse heaks tarneagentuuri 5. mai 1960. aasta eeskiri, millega nähakse ette, kuidas maakide, lähtematerjalide ja
        lõhustuvate erimaterjalide nõudlust pakkumisega tasakaalustada</seg>
      <seg lang="mt" n="3" part="N" TEIform="seg">li tiffissa d-data meta l-Aġenzija għall-Forniment tal-Euratom għandha tibda
        taqdi d-dmirijiet tagħha u li tapprova r-Regoli ta' l-Aġenzija tal-5 ta' Mejju 1960 li jistabbilixxu l-manjiera li fiha d-
        domanda għandha tiġi bilanċjata kontra l-provvista ta' minerali, materjali primi u materjali speċjali fissili</seg>
    </ab>
  - <ab type="1-1" part="N" TEIform="ab">
      <seg lang="et" n="4" part="N" TEIform="seg">EUROOPA ÜHENDUSTE KOMISJON,</seg>
      <seg lang="mt" n="4" part="N" TEIform="seg">IL-KUMMISSJONI TAL-KOMUNITÀ EWROPEA DWAR L-ENERĠIJA ATOMIKA,</seg>
    </ab>
  - <ab type="1-1" part="N" TEIform="ab">
      <seg lang="et" n="5" part="N" TEIform="seg">võttes arvesse Euroopa Aatomienergiaühenduse asutamislepingut, eriti selle
        artikleid 52, 53, 60 ja 222</seg>
      <seg lang="mt" n="5" part="N" TEIform="seg">Wara li kkunsidrat it-Trattat li jistabbilixxi l-Komunità Ewropea dwar l-Enerġija
        Atomika, u partikolarment l-Artikoli 52, 53, 60 u 222 tiegħu;</seg>
    </ab>
  - <ab type="2-1" part="N" TEIform="ab">
      <seg lang="et" n="6" part="N" TEIform="seg">ning arvestades, et:</seg>
      <seg lang="et" n="7" part="N" TEIform="seg">komisjon peab kinnitama kuupäeva, millal tarneagentuur alustab talle
        asutamislepinguga seatud ülesannete täitmist;</seg>
      <seg lang="mt" n="6" part="N" TEIform="seg">Billi hhija l-Kummissjoni li tiffissa d-data li fiha l-Aġenzija għall-Forniment
        għandha tidhol għad-dmirijiet li jgħaddu għandha taħt it-Trattat;</seg>
    </ab>
    ...
  </div>
```

Figure 2: Section of Estonian-Maltese sample alignment in XML format.

especially useful for legal texts: in the Acquis corpus, 6.5 percent of the tokens are numbers.

The one-to-one segments found in this first round of alignment are randomly sampled (10,000 sentence pairs in the case of the Acquis corpus) to feed the second phase of the algorithm: a simple automatic lexicon-building.

In the third phase the alignment is re-run, this time also considering similarity information based on the automatically constructed bilingual lexicon.

For a given language pair and the whole AC corpus, the running time of each of these three phases is about ten minutes on a fast personal computer. This makes it perfectly feasible to run the algorithm on all 210 language pairs of the corpus. We note that after incremental changes to the corpus, it is not necessary to re-run the first two phases.

## 5.3. Producing Parallel Corpora for all Language Pair Combinations

For the purpose of the distribution of the corpus, the paragraph alignment information from the database was exported into files (both XML and comma-separated values), one per language pair. In the corpus distribution, we provide a PERL script that can be used to generate a bilingual aligned corpus for any of the 190+ language pairs. The script reads the stand-off alignments (c.f. next section) for the language pair of interest, extracts the required paragraphs from the documents in the corpus and outputs them as in-place alignments. Figure 2 shows an Estonian-Maltese sample alignment.

## 6. Corpus Format

The corpus is encoded in XML, according to the Text Encoding Initiative Guidelines TEI P4 (Sperberg-McQueen & Burnard 2002). The corpus consists of two parts, the documents and the alignments.

The documents are grouped according to language; all the texts for one language constitute one TEI corpus, which consists of the TEI header, giving extensive information about the language corpus, and the actual documents. Each document contains, again, a TEI header, giving for instance the download URL and the EUROVOC codes, and the text, which consists of the title and a series of paragraphs.

The two-way alignments are, for each language pair, stored as a TEI document. However, the document does not contain actual texts, but only pointers to the aligned paragraphs. As explained above, these can with the help of the included program be converted into in-place alignments.

It should be noted that the headers are also available in HTML, and thus serve to introduce and document the corpus in the distribution.

## 7. EUROVOC Subject Domain Classification

Most CELEX documents have been manually classified into subject domain classes using the EUROVOC thesaurus (Eurovoc 1995), which exists in one-to-one translations into approximately twenty languages and distinguishes about 6,000 hierarchically organised descriptors (subject domains). Where available, we included the *numerical* EUROVOC codes into the header of the Acquis documents (see Figure 1). Users who are interested in seeing the descriptor *text* for each of the numerical descriptor codes should request a EUROVOC licence from opoce-info-copyright@cec.eu.int, mentioning the file reference number 2005-COP-395. To our knowledge, the licence is free of charge for research purposes and costs 500 Euro for commercial use.

The EUROVOC subject domain classification in combination with the JRC-Acquis can be used for at least two purposes: (1) the automatic generation of subject domain-specific monolingual or multilingual terminologies (e.g. Giguet & Luquet 2005) and (2) the training of automatic multi-label document classifiers and keyword indexing systems (e.g. Civera & Juan 2006; Pouliquen et al. 2003; Montejo Ráez 2006).

## 8. Summary

To our knowledge, the JRC Collection of the Acquis Communautaire – available currently in all twenty official EU languages plus Romanian – is the only parallel corpus of its size available in so many languages. The current version of the JRC-Acquis is distributed in TEI-compliant XML format. It is accompanied by paragraph segmentation and information on segment alignment using both Vanilla and HunAlign. It is furthermore accompanied by EUROVOC subject domain information for most texts. We hope that more meta-information will be added by the scientific community in the future and that the equivalent Croatian, Bulgarian and Turkish texts will be added, following the Romanian example. Please contact a JRC author if you have any useful information to add, especially if it is available for all languages. We would be happy to add this information, or a link to its location, to the JRC-Acquis web site. The JRC-Acquis corpus and related meta-information is available for download at http://wt.jrc.it/lt/acquis/.

## 9. Acknowledgements

We are grateful to the EC's *Office for Official Publications* OPOCE and to Mr. Alain Reichling from the EC's Translation Service in Luxembourg for their support and help in making this corpus available for research purposes.